\documentclass{article}
\usepackage{spconf,amsmath,graphicx}
\usepackage[nodisplayskipstretch]{setspace}
\usepackage{multirow}

\title{Pseudo Pixel-level Labeling for images with evolving content}
\name{Sara Mousavi$^{\star}$ \quad Zhenning Yang$^{\star}$ \quad Kelley Cross $^{\dagger}$  \quad Dawnie Steadman$^{\dagger}$ \quad Audris Mockus$^{\star}$}

\address{$^{\star}$  Department of Electrical Engineering and Computer Science\\
    $^{\dagger}$Department of Anthropology\\
    The University of Tennessee Knoxville, USA}

\begin{document}

\maketitle
\begin{abstract}
Annotating images for semantic segmentation requires intense manual labor and is a time-consuming and expensive task especially for domains with a scarcity of experts, such as Forensic Anthropology. We leverage the evolving nature of images depicting the decay process in human decomposition data to design a simple yet effective pseudo-pixel-level label generation technique to reduce the amount of effort for manual annotation of such images. 
We first identify sequences of images with a minimum variation that are most suitable to share the same or similar annotation using an unsupervised approach. 
Given one user-annotated image in each sequence, we propagate the annotation to the remaining images in the sequence by merging it with annotations produced by a state-of-the-art CAM-based pseudo label generation technique. 
To evaluate the quality of our pseudo-pixel-level labels, we train two semantic segmentation models with VGG and ResNet backbones on images labeled using our pseudo labeling method and those of a state-of-the-art method. 
The results indicate that using our pseudo-labels instead of those generated using the state-of-the-art method in the training process improves the mean-IoU and the frequency-weighted-IoU of the VGG and ResNet-based semantic segmentation models by $3.36$\%, $2.58$\%, $10.39$\%, and $12.91$\% respectively.
 

\end{abstract}
\begin{keywords}
Pseudo Labeling, Semi-supervised Semantic Segmentation, Evolving Data, Image Sequencing, Class Activation Map
\end{keywords}
\section{Introduction}
\label{sec:intro}
Pixel-level labeling for purposes such as instance and semantic segmentation is a time-consuming and costly endeavor. 
This problem is exacerbated when the annotation process cannot be crowd-sourced due to privacy concerns for the data at hand or when domain expert knowledge is required in the annotation process. 

Semi-supervised learning (SSL) has been used as a technique for increasing the performance of models when only a small portion of the data is fully annotated and the rest of the data is either weakly labeled or fully unlabeled \cite{ouali2020semi, lee2019ficklenet,zhou2019collaborative}. 
Adapted from SSL, pseudo-labeling techniques use hints from weakly labeled data to generate pseudo-pixel-level labeled data to be used for learning in a supervised manner.  
Many such methods are based on generating CAMs (Class Activation Maps) and refining them to generate high quality pseudo pixel-level labels \cite{zhou2018weakly,ahn2019weakly}.

In this work, we consider the problem of pseudo-labeling for images with evolving content. Some examples of such data are images depicting human-decomposition, produce decay, plant growth, human aging. 
These datasets contain subjects (various classes) that change over time with fixed or dynamic intervals often days or weeks apart. At any given timestep there exist one or more images depicting each class. 
In this paper, we utilize the underlying structure of such image data, specifically the similarity of discretely consecutive images of a subject, along with weak image-level labels to generate rich pseudo-pixel-level labels that can be used for training and improving the overall performance of a segmentation task.

Our proposed method in this work consists of three main steps.
First, we identify images of the same class with minimum variation that are most suitable to share the same or similar annotation and group them into sequences. 
A single image from each sequence is presented to a human annotator to be manually annotated. Second, we generate CAM-based pseudo-pixel-level labels for the available weakly labeled images \cite{ahn2019weakly, ouali2020semi}. Lastly, we merge the CAM-based pseudo-pixel-level labels with the single annotation provided by the annotator in order to generate pseudo-pixel-level labels for the remaining images in the sequences.

We evaluate our method using a dataset of human decomposition (described in Section \ref{sec:data}). We use VGG and ResNet-based semantic segmentation U-Nets and evaluate their performance when trained on images labeled using solely the CAM-based method and those from our proposed method. The results show that training on our pseudo-pixel-level labels improves both mean-IoU and the Frequency-weighted-IoU of the semantic segmentation task. 
This indicates that our method, although very simple, is an effective technique for generating pseudo-pixel-level labels for images with evolving content. Additionally, our proposed approach reduces the human experts' effort for image annotation from the number of images to the number of sequences in an unlabeled image dataset with evolving content. 

The rest of this paper is as follows. Related works are discussed in Section~\ref{sec:relatedwork}. Details of our method are included in Section~\ref{sec:method}. We discuss the evaluation of our method in Section \ref{sec:evaluation} which includes the decomposition data used in this work (Section~\ref{sec:data}) as well as the results (Section  \ref{subsec:ablation}). 
Finally, the paper is concluded in section \ref{sec:conclusion}.
\section{Related work} \label{sec:relatedwork}
\subsection{Manual Labeling}
With the growth of large data and novel computer vision techniques, the need for annotated data for various applications is growing. 
There are many tools that facilitate annotating images. For example, LabelMe~\cite{russell2008labelme} and similar polygon-drawing interfaces have been used to annotate image collections~\cite{Cordts_2016_CVPR}. 
In addition, other platforms such as Amazon Turk~\cite{daly2015swapping} are also widely used for crowdsourcing annotations. However, given the amount of time needed to annotate a single image which can take 15 to 60 times longer than that of region-level and image-level labels respectively~\cite{lin2014microsoft}, such approaches are not efficient nor always possible alone. Moreover, real-world image datasets in various disciplines may have privacy concerns or require human expertise, making it more difficult to rely on crowdsourcing. 

\subsection{Human-machine Labeling}
 Fluid Annotation~\cite{andriluka2018fluid} and other methods such as block annotation \cite{lin2019block} and the context-aware segmentation method presented by Majumder et l. \cite{majumder2019content} assist annotators in fully annotating images by providing initial annotations and allowing user interaction in the annotation process. 
 However, such approaches without first being trained on the target domain data perform poorly on unseen domain-specific datasets that lack adequate training data to begin with.

\subsection{Pseudo Labeling}\label{subsec:pseudo}
Other semi-supervised based approaches aim at improving the performance of computer vision tasks by leveraging the large amount of unlabeled or weakly labeled data. New efforts in this area take various measures, such as the use of pseudo-labels or consistency-based approaches, to take advantage of unlabeled samples. Wu and Prasad generated pseudo-labels from unlabeled data to train an initial model whose weights were used for a secondary classification network \cite{wu2017semi}. Other semi-supervised methods such as \cite{liu2019deep,iscen2019label} are based on label propagation to generate more training data. Ahn et al. propose IRNet and use CAMs and assign an instance label to each seed and propagate to find an accurate boundary around each instance. IRNet is trained on inner pixel relations on attention maps. In this work, we use a CAM-based technique to generate pseudo-pixel-level labels for images with evolving content and use the work done by Ahn et al. as our baseline. 
\section{Method} \label{sec:method}
The goal of our work is to reduce the human annotation effort in annotating datasets depicting evolving content by generating pseudo-pixel-level-labels for such images. Our method leverages the underlying structure of such datasets by utilizing the local similarity between images of an evolving subject within neighboring timesteps to generate sequences of similar images. We then adapt a single user-supplied annotation for each sequence to label other images in that sequence.

\begin{figure*}
    \centering
    \includegraphics[width=0.7\textwidth]{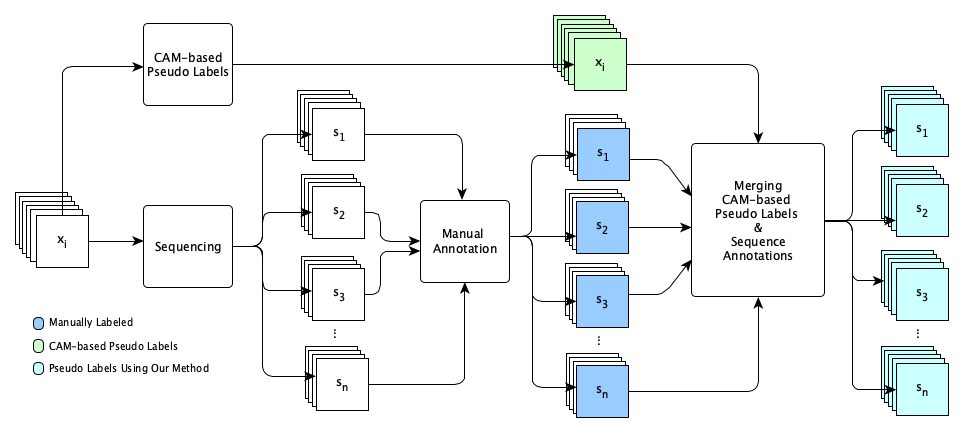}
    \caption{An overview of our proposed method. Unlabeled images ($x_i$) are fed into a CAM-based Pseudo label generator and a sequencing component to generate pseudo-pixel-level annotations and sequences of similar images respectively. Next, one image from each sequence is manually annotated. 
    Finally, the manually created annotation and the CAM-based pseudo-pixel-level labels are merged to form the final pseudo annotation.}
    \label{fig:overview}
\end{figure*}

Unlike the frames of a video, evolving image data do not always readily come in sequences with slight changes from one image to another. For example, in the human-decomposition data tackled in this work, various images are taken from different angles at varying intervals (e.g. multiple days apart) with each including one or more unlabeled body parts. 
Therefore, we first identify sequences of similar images in our data using an unsupervised approach \cite{mousavischism} and ask a domain user to annotate only one image for each sequence. We then merge the user-supplied annotation with CAM-based pseudo annotations generated for the other images of that sequence. The merging of the two annotations helps adjust for the variations in the subjects across the images of the sequences. 
Doing so, we eliminate the need for annotating every individual image and instead only one image will be manually annotated for each identified sequence. The overall structure of our method is shown in Figure \ref{fig:overview}.

\subsection{Image Sequencing}
Images of human decomposition include decaying subjects that evolve over time. This natural phenomena results in a decline in the level of similarity between images of the same subject from distant timesteps. For example, two images taken from the same area of the body at day $1$ and $2$ are more likely to be similar to each other than those from day one and day $10$. Due to such characteristics, we use a sequencing approach (SChISM) \cite{mousavischism} to generate sequences of similar images depicting the same area of the body. A domain expert human annotator is then presented with one image from each sequence for manual annotation. We denote these sequences with $\left \{ s_1 = \left \{ x_1, x_2, .., x_k \right \},..., s_n= \left \{ x_{m}, x_{m+1},..., x_{I}  \right \}\right \}$, 
where $n$ is the total number of generated sequences, and $I$ is the total number of images. 
We refer to the annotation manually obtained for the $i$th image in sequence $s$ as $y_{s}^{i}$. Each annotation includes classes ($C_s$) and the background ($B_s$).  

\subsection{CAM-based Pseudo Labeling}\label{subsec:cam}
The single user-supplied annotation $y_{s}^{i}$ is not suitable for other images on its own as images in the same sequence, while similar in content, can be different in terms of orientation, camera angle, level of decay, etc. In order to adapt the manual annotation to other images in the same sequence, we merge them with CAM-based pseudo-pixel-level labels.

The CAM-based pseudo-pixel-level label generation method used in this work is a state of the art pseudo pixel-level label generation technique presented by Ahn et al. \cite{ahn2019weakly} in CVPR 2019 for instance segmentation and adapted by Quali et al \cite{ouali2020semi} in CVPR 2020 for semantic segmentation. In this method, pseudo labels are generated by adding a classification branch to a pre-trained encoder, similar to previous works  \cite{ahn2018learning, lee2019ficklenet}, 
to generate pseudo-pixel-level labels. First, CAMs are generated using a method based on global average pooling presented by Zhou et al \cite{zhou2016learning}. Next, the pseudo pixel-level labels are generated using a background and a foreground threshold. Pixels with an attention score greater than the foreground threshold are considered as foreground, those with an attention score less than the background threshold are considered as background, and the rest of the pixels are ignored. The result is then refined using dense CRF \cite{krahenbuhl2011efficient}. 

We refer to each input image as $x_i$ and its corresponding CAM-based pseudo-pixel-level label as $y_{p}^{i}$. Each pseudo-pixel-level label $y_{p}^{i}$ contains pixel values depicting foreground, $F_p$ and discriminative areas within the $F_p$ for each class, $C_p$, which assigns the class index to its corresponding pixels, and background, $B_p$.

\subsection{Merging CAM Labels with Sequence Labels}
As a final step of our pseudo-label generation, we merge the manually generated annotations for each sequence with the CAM-based pseudo-pixel-level labels. This is done to use the discriminative pixels it has detected as a guide for the pseudo-pixel-level label generation for the rest of the images in the sequences. Essentially, the final pseudo annotation is generated by merging $y_{s}^{i}$ with $y_{p}^{i}$. We denote the resulting pseudo pixel-level label generated from this merge as $y_{ps}^{i}$.
To do so, we start by setting $y_{ps}^{i} = y_{s}^{i}$ for the $i$th image in the $s$th sequence. 
We then update each pixel of $y_{ps}^{i}$ by comparing its value to the pixel value at the corresponding position from $y_{s}^{i}$. 
For pixels that are only assigned to a class in $y_{p}^{i}$ and not in $y_{s}^{i}$, we use the pixels values of $y_{p}^{i}$ as shown in Equation \ref{equ:merge}.

\begin{align}\label{equ:merge}
\left\{\begin{matrix}
 y_{ps}^{i}[j] =  y_{p}^{i}[j]& j\in C_p\ and\ j \notin C_s \\ 
 y_{ps}^{i}[j] =  y_{s}^{i}[j] & Otherwise
\end{matrix}\right.
\end{align}
\vspace{-0.7cm}

\begin{table*}[]
\small
\centering
\begin{tabular}{|c|c|c|c|c|c|c|}
\hline
\multirow{2}{*}{} & \multicolumn{2}{c|}{Data} & \multicolumn{2}{c|}{Mean IoU (\%)} & \multicolumn{2}{c|}{Frequency weighted IoU (\%)} \\ \cline{2-7} 
 & \# Manual Labels & \# Pseudo Labels & \begin{tabular}[c]{@{}c@{}}VGG-16 \\ based U-Net\end{tabular} & \begin{tabular}[c]{@{}c@{}}ResNet-50\\ based U-Net\end{tabular} & \begin{tabular}[c]{@{}c@{}}VGG-16\\ based U-Net\end{tabular} & \begin{tabular}[c]{@{}c@{}}ResNet-50\\ based U-Net\end{tabular} \\ \hline
Supervised & 1036 & 0 & 32.43 & 33.88 & 56.98 & 60.17 \\ \hline
Weakly Supervised CAM & 518 & 518 & 21.94        & 25.81         & 46.63      & 48.98  \\ \hline
Weakly Supervised Our & 518 & 518 & 28.94        & 32.29         & 54.39       & 56.74\\ \hline

 \multirow{2}{*}{\begin{tabular}[c]{@{}c@{}}Weakly Supervised\\ CAM\end{tabular}} & 0 & 5965 & 26.27 & 20.01 & 45.44 & 41.04 \\ \cline{2-7} 
 & 1036 & 5965 & 36.88 & 34.08 & 60.82 & 43.43 \\ \hline
\multirow{2}{*}{\begin{tabular}[c]{@{}c@{}}Weakly Supervised\\ Our\end{tabular}} & 0 & 5965 & 29.63 & 30.40 & 48.02 & 53.95 \\ \cline{2-7} 
 & 1036 & 5965 & \textbf{37.66} & \textbf{37.79} & \textbf{61.99} & \textbf{61.66} \\ \hline 
 
\end{tabular}
\caption{Comparing the performance of semantic segmentation when training the models on the pseudo-pixel-level labels generated using our method and the baseline method.}
\label{tab:results}
\end{table*}

\section{Evaluation} \label{sec:evaluation}
In order to evaluate our proposed method, we use a semantic segmentation U-Net with two backbones of VGG and ResNet. We train the networks on various combinations of pseudo and actual labeled images through an ablation study explained in Section \ref{subsec:ablation}. As our baseline pseudo-labels, we use the CAM-based pseudo-labels generated following the procedure described in Section \ref{subsec:cam}. For our evaluation, we use a total of $7445$ images, $1480$ of which are manually annotated and the remaining $5965$ images are only image-level-labeled. From the $1480$ manually annotated images, $444$ images are randomly selected as our test data. 
We report mean-IoU and frequency weighted IoU that are calculated as $\frac{1}{n_{cl}}.\frac{\sum_{i}n_{ii}}{t_i + \sum_{j}n_{ji}-n_{ii}}$ and $\frac{1}{n_{cl}}.\frac{\sum_{i} t_in_{ii}}{(\sum_{k} t_{k}).(t_i + \sum_{j}n_{ji}-n_{ii})}$ respectively where $n_{ij}$ is the number of pixels of class $i$ predicted to belong to class $j$, where there are $n_{cl}$ different classes, and $t_i = \sum_{j} n_{ij}$ and $t_k$ are the total number of pixels (FN + TP) of class $i$ and all classes respectively.


\subsection{Human Decomposition Dataset}\label{sec:data}
The human decomposition dataset used in this work consists of photos taken daily of different body parts from various subjects donated to the Forensic Anthropology Center of our university. The subjects were placed in an uncontrolled environment 
to record and track the decay process. The dataset includes more than one million images and has been collected over 9 years, with each subject staying in the facility for approximately one year. At each session, multiple images depicting various body parts from different angles for each subject have been taken. 
Each body part represents one class. In this work, we consider $6$ main classes for ``hand'', ``arm'',  ``foot'', ``leg'', ``torso'', and ``head''. 

\subsection{Ablation Study}\label{subsec:ablation}
\begin{figure}
    \centering
    \includegraphics[height=.27\textwidth]{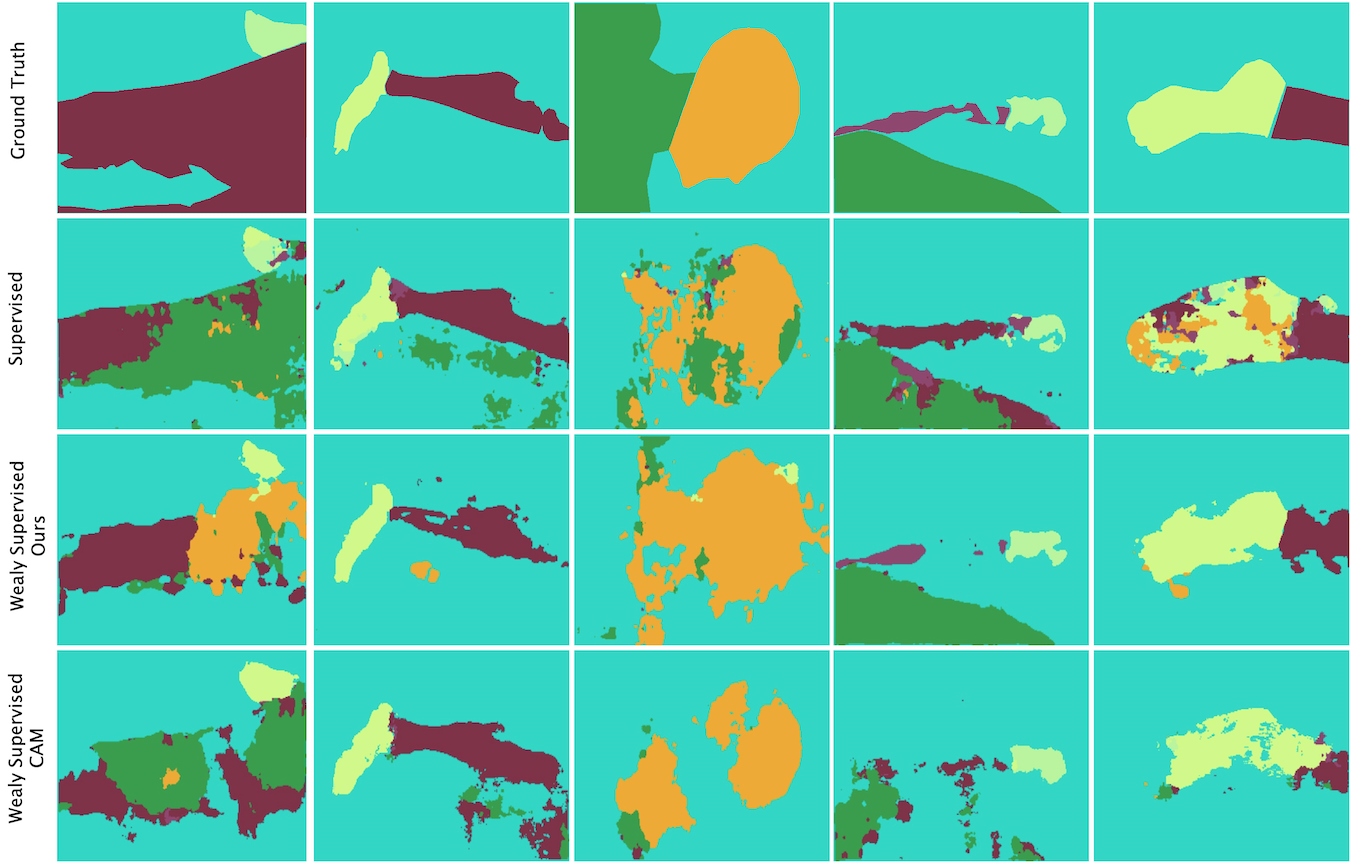}
    \caption{Comparing the semantic segmentation performed by a supervised and two weakly supervised models on a few images from the test set. In the supervised approach the model is trained on only the $1036$ manually labeled images where as in the weakly supervised approaches the models are trained on $1036$ manually labeled plus $5965$ pseudo-labeled images. }
    \label{fig:example}
\end{figure}

In order to assess 
the quality of the pseudo labels and their impact on the overall learning process, we conduct the following ablation study. We first consider a supervised scenario where we only have $1036$ manually labeled images depicting $6$ classes in the training process and consider this as a baseline without any pseudo labels. Next, we consider the performance of the models, once when they are trained on the CAM-based pseudo labeled images and once with the addition of the manually labeled images. 
Finally, we followed the same procedure for image labels produced by our pseudo labeling method (once with and once without the manually labeled images). Table \ref{tab:results} shows the results for these experiments. Furthermore, an example comparing the predictions on a few images from the test set is shown in Figure \ref{fig:example}. 

In most cases, the VGG-based U-Net produces better and more consistent results. 
The deeper the model the more sensitive it is to noise, and it converges much slower especially when training it with pseudo labeled data.
Our proposed method, further refines the pseudo labels at a pixel level by ensuring the consistency between the manually-provided labels for the sequences and CAM-based labels.
Training VGG and ResNet with only our refined pseudo labels increases the mean IoU and frequency-weighted-IoU by $3.36$\%, $2.58$\%, $10.39$\%, and $12.91$\% respectively. Furthermore, the segmentation in Figure \ref{fig:example} indicates our method results in better boundaries for the objects depicted in the images.

\section{Conclusion}\label{sec:conclusion}
In this work, we aim at reducing the number of images needed to be manually annotated by a domain expert for training a semantic segmentation model on images with evolving content. We propose a method that leverages the underlying structure of such datasets to generate pseudo labels for the images when only a small portion of them are manually annotated. Our method first identifies a set of representative images to be suggested to a human annotator for manual annotation and then propagates these annotations to the rest of the images by merging them with CAM-based annotations generated using the IRNet method \cite{ahn2019weakly}. In the future, we plan on extending this approach by incorporating spatial alignment methods to better adapt the single user-provided annotation to the remaining images in the sequences. 

\clearpage
\bibliographystyle{IEEEbib}
\bibliography{refs}

\end{document}